\documentclass[conference]{IEEEtran}
\IEEEoverridecommandlockouts
\usepackage{cite}
\usepackage{amsmath,amssymb,amsfonts}
\usepackage{algorithmic}
\usepackage{graphicx}
\usepackage{textcomp}
\usepackage{xcolor}
\usepackage{fancyhdr}
\def\BibTeX{{\rm B\kern-.05em{\sc i\kern-.025em b}\kern-.08em
    T\kern-.1667em\lower.7ex\hbox{E}\kern-.125emX}}
    
\fancypagestyle{FirstPage}{

\lfoot{978-1-6654-0575-1/22/\$31.00 \copyright2022 IEEE}

}

\begin{document}

\title{MoRSE: Deep Learning-based Arm Gesture Recognition for Search and Rescue Operations\\
}

\author{\IEEEauthorblockN{Panagiotis Kasnesis}
\IEEEauthorblockA{\textit{Electrical and Electronics Engineering} \\
\textit{University of West Attica}\\
Egaleo, Greece \\
pkasnesis@uniwa.gr}
\and
\IEEEauthorblockN{Christos Chatzigeorgiou}
\IEEEauthorblockA{\textit{Electrical and Electronics Engineering} \\
\textit{University of West Attica}\\
Egaleo, Greece \\
chrihatz@uniwa.gr}
\and
\IEEEauthorblockN{Dimitrios G. Kogias}
\IEEEauthorblockA{\textit{Electrical and Electronics Engineering} \\
\textit{University of West Attica}\\
Egaleo, Greece \\
dimikog@uniwa.gr}
\and
\IEEEauthorblockN{Charalampos Z. Patrikakis}
\IEEEauthorblockA{\textit{Electrical and Electronics Engineering} \\
\textit{University of West Attica}\\
Egaleo, Greece \\
bpatr@uniwa.gr}
\and
\IEEEauthorblockN{Harris V. Georgiou}
\IEEEauthorblockA{\textit{Hellenic Rescue Team of Attica (HRTA)} \\
Athens, Greece \\
harris@xgeorgio.info}
\and
\IEEEauthorblockN{Aspasia Tzeletopoulou}
\IEEEauthorblockA{\textit{Hellenic Rescue Team of Attica (HRTA)} \\
Athens, Greece \\
atzelet@gmail.com}

}

\maketitle

\begin{abstract}
Efficient and quick remote communication in search and rescue operations can be life-saving for the first responders. However, while operating on the field means of communication based on text, image and audio are not suitable for several disaster scenarios.
In this paper, we present a smartwatch-based application, which utilizes a Deep Learning (DL) model, to recognize a set of predefined arm gestures, maps them into Morse code via vibrations enabling remote communication amongst first responders. The model performance was evaluated by training it using 4,200 gestures performed by 7 subjects (cross-validation) wearing a smartwatch on their dominant arm. Our DL model relies on convolutional pooling and surpasses the performance of existing DL approaches and common machine learning classifiers, obtaining gesture recognition accuracy above 95\%. We conclude by discussing the results and providing future directions.
\end{abstract}

\begin{IEEEkeywords}
deep learning, wearable computing, gesture recognition, search and rescue
\end{IEEEkeywords}

\section{Introduction}
\thispagestyle{FirstPage}

Wrist-worn devices (e.g., smartwatches and wristbands) are widely used electronics for recording data, such as motion signals produced by embedded motion sensors (e.g., accelerometers, gyroscopes, etc.). The wearable market was valued at 35.36 billion EUROS in 2020 and is expected to expand at a compound annual growth rate (CAGR) of 13.8\% from 2021 to 2028 \cite{market}. This is mainly due to their properties of being unobtrusive means of communication and capable of monitoring their users' vital signs \cite{Reiss2019DeepPL,Burrello2021QPPGEP} (e.g., heart rate), emotions \cite{Schmidt2018IntroducingWA}, daily activities \cite{Stisen2015SmartDA}, etc. 

Over the last few years, common Machine Learning (ML) and Deep Learning (DL) techniques have been applied broadly and successfully to recognize the gestures and activities performed by users wearing electronic devices, thus, they are exploited for effective gesture-based human-machine interaction \cite{Zhu2018ControlWG}. In particular, ML/DL applications have been used to several domains, such as smart homes \cite{Laput2016ViBandHB} to recognize gestures and send the corresponding commands to the home automation platform \cite{Luna2017WristPA,Nascimento2019NetflixCM}, smart factories assisting workers in performing preventive and corrective machine maintenance \cite{Villani2016SmartwatchEnhancedIW}, gesture-based UAV control \cite{Choi2017WearableGC}, automotive \cite{Maecki2020GestureBasedUI}, even empowering physical security \cite{kasnesis2019gesture}. 

Regarding physical security, arm signals are internationally used in several field operations, ranging from military services, firefighters, coast guards, lifeguards, police, medical First Responders (FRs), etc. They are, also, used in various working sectors for specialized operations such as cranes, loading and unloading of trucks to manage speed, direction and safety procedures. During the basic training courses, FRs learn different ways to communicate on the field. These include audio signals, arm/hand gestures, RF communications, visual signals and in some domains even flag signalling. In case of emergency or degraded communications (e.g. underground) or in extremely noisy environments (e.g. using heavy machinery during victim extrication), FRs on Search and Rescue (SaR) operations (e.g., emergency medical services or emergency medical technicians) should employ a common approach for gesture communication \cite{usar}. However, in these cases at least one member of the SaR team should keep the FR performing the gesture in view.

In the current paper we propose MoRSE (arM Movement Recognition for firSt rEsponders), which is an application that is deployed in the smartwatches of the FRs in SaR operations to capture and identify arm gestures exploiting a deep Convolutional Neural Network (CNN). In particular, 5 Gestures of Interest (GoI) have been selected, that, if a FR performs one of them a non-visual/non-audible message, is sent to the corresponding FR member of his/her team, through vibrations on his/her smartwatch device. To this end, we built MoRSE dataset to train the DL model and deploy it, afterwards, to smartwatches, to classify in real-time incoming motion signals to predefined gestures. 

In the next section related works regarding wearable-based gesture recognition for improving public safety are presented. Section 3 describes the developed application, providing explanation on selecting the 5 GoI and introduces the architecture of the selected algorithm. The collected MoRSE dataset is described in section 4 and demonstrates the obtained results. Finally, section 5 concludes the paper discussing future steps.

\section{Related works}

In the current section, we report mobile/wearable approaches that have been proposed in the literature for empowering public safety and are based on machine learning (ML) techniques. Moreover, these works exploit Inertial Measurement Unit (IMU) sensors to classify the arm/hand gestures and the activities  explicitly (i.e., the users perform them intentionally for notification purposes), or implicitly where the system identifies the situation they are in.

Car accident detection based on motion sensor data is an ML task that has been explored in several works \cite{ferreira2017localization}\cite{megalingam2010wireless}\cite{goel2018detecting}. In \cite{ferreira2017localization} the authors exploit the smartphone sensors to provide situational awareness to emergency responders; they rely on GPS and accelerometer sensors to resist false positives and discuss how smartphone-based accident detection reduces traffic congestion and increases the preparedness of emergency responders. On the other hand, Goel et al. \cite{goel2018detecting} try to detect the driver’s distraction into order to prevent accidents and increase road safety. In their experiment, they achieved promising results using a driving simulator and 16 adult subjects that were asked to wear on their wrist a smart device equipped with an accelerometer and a gyroscope. Similar work was done by Tavacoli et al. \cite{Tavakoli2021DriverSA}, however, the authors exploited more sensor modalities (e.g., Photoplethysmogram) and executed their experiments in the wild.  In addition to this, distraction has been evaluated, also, from the pedestrian perspective \cite{vinayaga2018towards}, where a novel complex activity recognition framework called DFAM was designed exploiting motion data from the users’ mobile and wearable devices and evaluated against several baseline classifiers (i.e., k-NN,  random forests, decision trees, naïve Bayes, support vector machines).

Gunshot recognition has been explored in \cite{loeffler2014detecting}\cite{khan2018firearm}. In \cite{loeffler2014detecting} logistic regression is used to identify patterns on wrist movements to detect gunshots based on wearable sensors data; it obtained 99.4\% in terms of classification accuracy when tested against a hold-out sample of observations. A more complex gunshot detection task is presented in \cite{khan2018firearm}, where the machine learning (ML) algorithm should identify various categories of firearms (i.e., handgun, rifle, shotgun) and recognize whether a firearm is autoloaded or manual. The best results were obtained using a decision trees classifier.

Fall detection based on mobile or wearable devices is also a well-investigated computer task \cite{figueiredo2016exploring}\cite{maglogiannis2014fall}. Fall detection is considered to be crucial in the case of the elderly people, since falls may lead to their hospitalization and sometimes are fatal, thus, fast assistance is important. In \cite{figueiredo2016exploring} the authors used a smartphone equipped with an accelerometer, magnetometer and gyroscope sensors and obtained up to 93\% for specificity and 100\% for sensitivity. The work introduced in \cite{maglogiannis2014fall} focuses on reducing false alarms (i.e., false positives) using the accelerometer data of a smartwatch. 


Finally, incident reporting through the use of gestures is reported in \cite{kasnesis2019gesture}. The authors proposed and evaluated an incident reporting application that can be deployed on smartwatches and uses a late sensor fusion deep CNN model to recognise gestures. This technique allows citizens to report hazardous events using discreet arm signals to alert the law enforcement agents. The proposed approach was evaluated using a dataset the authors build and managed to produce accuracy above 98\% and F1-score over 97\%.

\section{MoRSE application}
\begin{figure}[t]
\centerline{\includegraphics[width=\columnwidth]{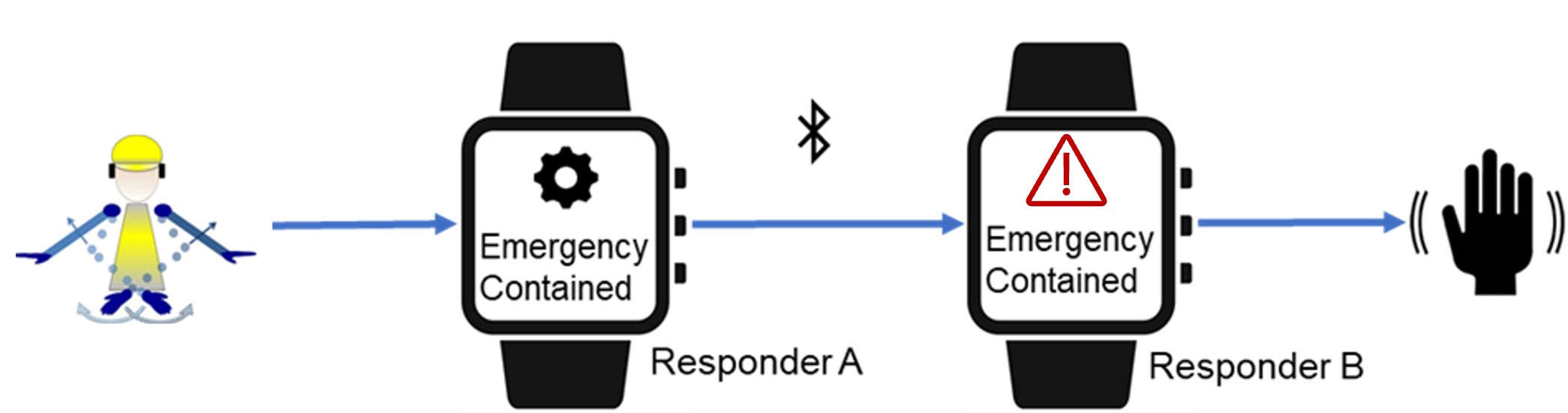}}
\caption{MoRSE data processing and communication pipeline. Arm gesture is performed by a FR (Responder A) it is recognized at his/her smartwatch, and transmitted to his/her colleague's (Responder B) smartwatches and translated into Morse code using vibrations.
\label{fig0}}
\end{figure}

In the current section we introduce MoRSE application, describing in detail the selected GoI, the examined DL and ML algorithms used for recognizing them and, finally, the communication pipeline and mapping to Morse code (Figure~\ref{fig0}).  

\subsection{Selected hand signals}

For our solution, four widely used arm/hand signals taken from airport ground control operations \cite{skybrary} were selected as emergency arm signals to be used in such an automated procedure. They are widely known and easy to perform for every FR on the field, almost under any condition. Additionally, a fifth hand signal was developed to indicate an FR in distress/emergency. 

These five signals cover the situations of:

\begin{figure*}[htbp]
\centering\includegraphics[width=\textwidth]{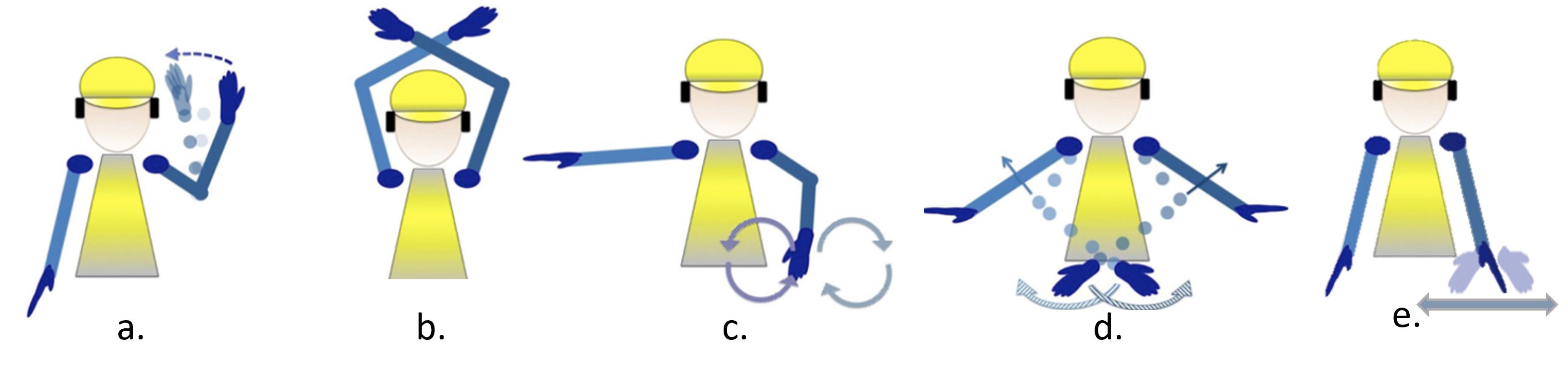}
\caption{Illustrations of the 5 selected Emergency Arm Signals \cite{skybrary}.
\label{fig1}}
\end{figure*}

\begin{enumerate}

\item \textbf{Recommended Evacuation (RE)}. A back and forth movement of the arm indicates emergency evacuation (Figure~\ref{fig1}b).

\item \textbf{Recommended Stop (RS)}. An “X” is made with the hands above the head indicates stop activities or stop works in progress, including evacuation, if in progress (Figure~\ref{fig1}b). 

\item \textbf{Fire (F)} A movement in the shape of number 8 performed with the hand that the smartwatch is on indicates fire alert (Figure~\ref{fig1}c).

\item \textbf{Emergency Contained (EC)}. It means that activities can be resumed indicating “all clear” and it is shown with arms extended outwards and down until wrist crossed and backwards at 45-degree angle (Figure~\ref{fig1}d).

\item \textbf{Distress (DS)}. The last signal is repeatedly rotating movements of the wrist (leftwards and rightwards) and indicates a FR in distress. This movement is chosen because it can be performed in emergency situations in very confined spaces which is a common place for an FR to be, e.g. inside narrow tunnels, vertical ascend/descend in elevator shafts, crawling under semi-collapsed structures, etc. (Figure~\ref{fig1}e).

\end{enumerate}

The rationale behind the selection of these five hand signals is based on the fact that they differ from each other significantly in timing and spatial movement, thus it is difficult to be misinterpreted. Furthermore, they can be performed in places where there is limited space, or the body position of the FR permits only limited movement.  During a massive disaster event, the number of FRs operating on the field varies from a few tenths to hundreds, from different operational fields (e.g., police, USAR, medics, civil protection) and perhaps different countries. Hence, arm signals must be as clear as possible and at the same time easy for FRs to learn and perform. It has been shown that arm signals used internationally in airport ground control are ideal candidates for such use. Regarding the fifth arm signal (distress), it is expected that FR many times find themselves in need of immediate assistance. Thus, such an arm signal that can be easily performed in very confined spaces can be lifesaving. Finally, we should mentioned that an extra movement called "random" was recorded including a high variance of arm movements not belonging to any of the aforementioned classes (see section IV).

\subsection{Selected algorithms}

The proposed network, is a lightweight deep CNN in order to be deployed on the smartwatch (i.e., contains around 55,000 parameters). Its architecture is very similar to previously developed deep CNN \cite{kasnesis2018perceptionnet}, and consists of the following layers: 
\begin{itemize}
    \item \textit{layer 1}: 12 convolutional filters with a size of (1,11), i.e., W\textsubscript{1} has shape (1, 11, 1, 12). This is followed by a ReLU activation function, a (1,4) strided pooling operation and a dropout probability equal to 0.5.
    \item \textit{layer 2}: 24 convolutional filters with a size of (1,11), i.e., W\textsubscript{2} has shape (1, 11, 12, 24). Similar to the first layer, this is followed by a ReLU activation function, a (1,2) strided pooling operation and a dropout probability equal to 0.5.
    \item \textit{layer 3}: 32 convolutional filters with a size of (6,11), i.e., W\textsubscript{3} has shape (6, 11, 24, 32). The 2D convolution operation is followed by a ReLU activation function, a 2D global pooling operation and a dropout probability equal to 0.5.
    \item \textit{layer 4}: 32 hidden units, i.e., W\textsubscript{4} has shape (32,6), followed by a softmax activation function.
\end{itemize}

Moreover, starting from the fact that a neural network can represent any function provided it has sufficient capacity (i.e. sufficiently broad and deep to represent the function), we examined, also, convolutional pooling (i.e., as a pooling operation apart from max pooling) which is a biologically inspired pooling layer modelled on strided convolutional operations \cite{Shen2014ALS}. It can be imagined as the convolutional pooling gives increased weight to stronger features and suppresses weaker features, similar to a latent space, thus, we call it \textit{latent pooling} operation; the designed deep CNN that utilizes them is denoted from now on as CNN-lp.

The proposed deep CNN algorithms were evaluated by comparing their performance to Convolutional LSTM \cite{ordonez2016deep} and conventional classifiers: Logistic Regression (LR), k-Nearest Neighbours (k-NN), Decision Tree (DT), and Random Forest (RF). We selected to extract only seven time-dependent features for each axis of the accelerometer and gyroscope signals, resulting into 42 features in total, since, they have been proven to be more robust and efficient than frequency-dependent features in activity recognition \cite{figo2010preprocessing}. Moreover, using too many features in some cases decrease the accuracy \cite{goel2018detecting} and result to overfitting. The selected features over all the sensor modalities are: a) mean value, b) minimum value, c) maximum value, d) median value, e) standard deviation, f) skewness (i.e., degree of asymmetry of the signal distribution), and g) kurtosis (i.e., degree of peakedness of the signal distribution).


Finally, the Convolutional LSTM architecture is very similar to that presented in \cite{ordonez2016deep}. In particular, the network consists of 4 convolutional layers each containing 32 kernels with filter size (1,5) and their output is processed by an LSTM layer. This network contains around 29,000 parameters.

\subsection{Mobile application}
The MoRSE application has been mainly developed for smartwatches, however, it communicates also with the paired smartphone to act as a gateway to the network in case the detected event (i.e., recognized gesture) should be sent to a central control of operation platform.


In particular, the smartwatch features include:
\begin{itemize}
    \item Gesture recognition (uses the DL model for inference)
    \item Location acquisition (exploits the GPS module)
    \item Gesture scanning from nearby FRs (utilizes Bluetooth Low Energy protocol)
    \item Translation of detected gesture to Morse code through vibrations
\end{itemize}


The gesture recognition features consumes the motion data signals produced by the embedded 3axial accelerometer and gyroscope sensors.
Every second the collected signals are processed by the deployed DL model and the result (captured gesture) is displayed on the smartwatch Messages screen. 

\begin{figure}[htbp]
\centerline{\includegraphics[width=4cm]{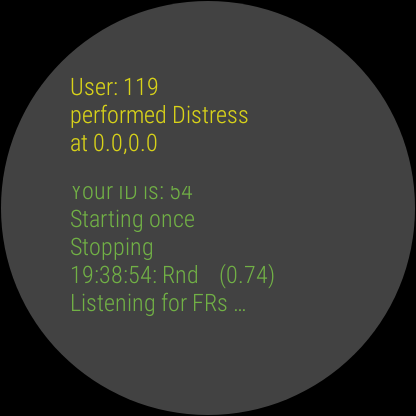}}
\caption{Messages screen in smartwatch.\label{fig9}}
\end{figure}

In particular, the Messages screen is divided horizontally into two halves (Figure~\ref{fig9}). The upper half displays (with yellow letters) information about any received message. Information includes the user ID, the gesture the user performed and their location (if available). The lower half displays (with green letters) the current user’s ID, the operation mode (transmitting, recognizing) and the last recognized gesture along with the relative time and the confidence of the recognition. 

Moreover, when a gesture is detected, it is broadcasted for 10 seconds to nearby FRs using Bluetooth Low Energy (BLE), while the users can also scan messages broadcasted from nearby FRs manually (exploiting BLE) by pressing the corresponding button. 
Message transmission is performed if all of the following are true:
\begin{itemize}
    \item A gesture is not Rnd (Random)
    \item Another gesture is not currently being transmitted
    \item The captured gesture is different than the previous.
\end{itemize}
The user will be notified with a notification in the smartwatch stating that transmission has been initiated. If the device is going to be operated in a closed space (e.g., building), location tracking has to be disabled in the application's settings. Location information is acquired from the GPS, and gesture transmission happens after the device gets the location. However, in closed spaces, location acquisition is almost always impossible since there is no line of sight with the GPS satellites. Therefore, the device will never get the location and will never transmit the gesture. 




When a message is received from nearby FRs, the device vibrates several times depending on the gesture pattern as shown in Table~\ref{tab4}. This is a a very important functionality enabling a wide range of messages via Morse code, which is a hands-free, non-verbal way of communication, provided that the FR is trained on it.
In particular, each "." symbol is translated into a 200ms vibration, the "-" symbol into 400ms vibration, while the space (i.e., no symbol) is mapped into 400ms of no vibration.

\begin{table}[htbp] 
\caption{Vibration patterns depending on the recognized gesture.\label{tab4}}
\begin{center}
\begin{tabular}{|c|c|}
\hline
\textbf{Gesture}	& \textbf{Morse code}\\
\hline
Recommended Stop		& .-. ...\\
Emergency Contained	        & . -.-.\\
Recommended Evacuation         & .-. .\\
Fire      & ..-.\\
Distress  & -.. ...\\
\hline
\end{tabular}
\end{center}
\end{table}



\section{Results}
In this section we present the collected MoRSE dataset, the experimental results and provide a discussion on them.

\subsection{Dataset}
For the purpose of this research, a dataset that includes the aforementioned Gestures of Interest (GoI) performed by 5 subjects has been collected. In particular, the participants were asked to wear a FOSSIL GEN 5 smartwatch, on the wrist of the arm they performed the moves and annotated them using the MoRSE data collection/annotation mobile application. Moreover, the subjects were asked to perform, except from the 5 GoI, some everyday life movements by acting naturally (e.g., do computer work, walking, scratching head etc.) while they wore the device, constituting the sixth class (denoted as random) in our classification scheme.

The MoRSE dataset consists of GoI performed by 7 subjects; contains around 4,200 data samples (approximately 600 for each class), which are 5-second segments including data values from the 3-axial gyroscopes and accelerometers of the smartwatch. It should be noted that the average sampling rate for both motion sensors is approximately 50 Hz, thus, each segment contains 250 values for each modality. In order to obtain subject dependent results, we used a 7-fold cross-validation technique and split our dataset into a training set containing 5 subjects, a validation set and a test set containing 1 subject. Finally, each sensor signal’s values were normalized by subtracting the mean value and dividing by the standard deviation. Table~\ref{tab3} presents the distribution of the collected gestures per subject.

\begin{table}[htbp] 
\caption{Distribution of the performed gestures per subject.\label{tab3}}
\begin{center}
\begin{tabular}{|c|c|c|c|c|c|c|c|}
\hline
\textbf{Subject No} & \textbf{Hand} & \textbf{Rnd}& \textbf{SA} & \textbf{RA} & \textbf{E} & \textbf{Fire} & \textbf{Distress}\\
\hline
1   & Left  & 100   & 100   & 100   & 100   & 100   & 101\\
2   & Right  & 100   & 100   & 100   & 100   & 100   & 100\\
3   & Right  & 100   & 71   & 101   & 99   & 99   & 100\\
4   & Right  & 100   & 100   & 100   & 100   & 100   & 100\\
5   & Right  & 100   & 100   & 101   & 101   & 106   & 115\\
6   & Left  & 100   & 103   & 102   & 102   & 101   & 101\\
7   & Left  & 100   & 100   & 100   & 100   & 100   & 100\\
\hline
\end{tabular}
\end{center}
\end{table}

\subsection{Experimental results}
The ML experiments were executed on a computer workstation equipped with an NVIDIA GTX 1080Ti GPU, which has 11 gigabytes RAM, 3584 CUDA cores, and a bandwidth of 484 GB/s. Python was used as programming language, and specifically the Numpy for matrix multiplications, data pre-processing, segmentation and the Tensorflow 2.0 framework for developing the neural networks. In order to accelerate the tensor multiplications, CUDA Toolkit in support with the cuDNN was used, which is the NVIDIA GPU-accelerated library for deep neural networks. The software is installed on a 18.04 Ubuntu Linux operating system.

The model was trained using the Adam optimizer \cite{kingma2014adam} with the following hyper-parameters: learning rate =0.001, beta1 = 0.9, beta2 = 0.999, epsilon=1e-08, decay=0.0. Moreover, we set the minimum number of epochs to 2000; however, the training procedure terminated automatically whether the best training accuracy had improved or not, after a threshold of 200 epochs. The training epoch that achieved the lowest error rate on the validation set was saved, and its filters were used to obtain the accuracy of the model on the test set.

\begin{figure}[htbp]
\centerline{\includegraphics[width=\columnwidth]{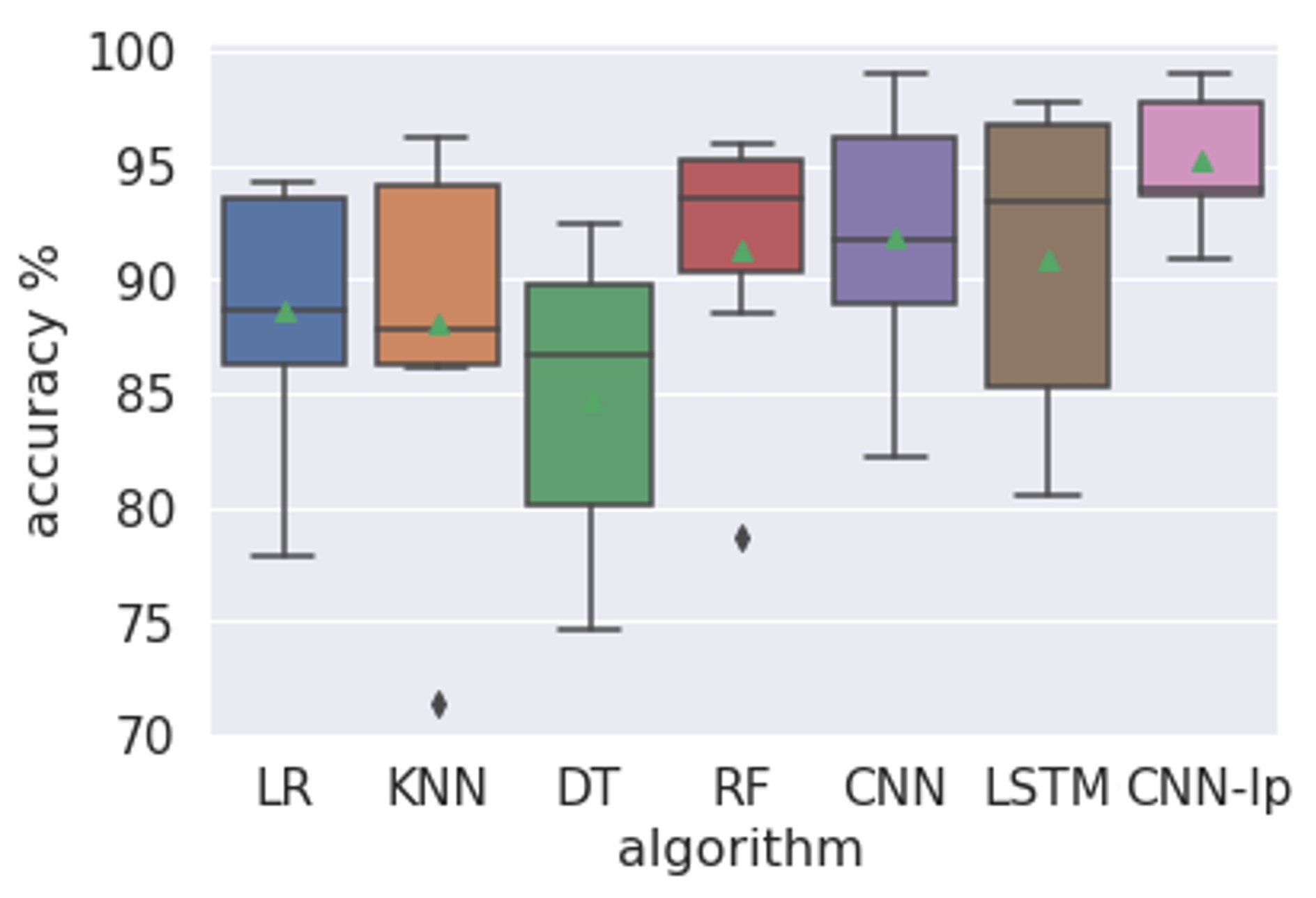}}
\caption{The performance, in terms of accuracy, of the developed machine learning algorithms.
\label{fig6}}
\end{figure}

Table~\ref{tab2} presents the accuracy results that were obtained by applying the aforementioned algorithms to our dataset. The best accuracy was achieved by the developed deep CNN-lp model (95.27\%), which utilizes the latent pooling layer and surpassed (by a lot) the developed architecture relying on max pooling (i.e., deep CNN). It is, also, worth mentioning that the presented results are the average values of the 5 runs that were executed for each fold. In addition to this, Figure~\ref{fig6} shows the box plots of the obtained results, with KNN having the biggest deviation in terms of accuracy among the 7 subjects and Deep CNN-lp the smallest one, making it more robust and reliable in real-world applications.

\begin{table}[htbp] 
\caption{Performance of each ML model on the MoRSE dataset.\label{tab2}}
\begin{center}
\begin{tabular}{|c|c|}
\hline
\textbf{Method Name}	& \textbf{Accuracy}\\
\hline
LR		& 88.59\%\\
KNN	        & 88.03\%\\
DT         & 84.78\%\\
RF      & 91.34\%\\
Convolutional LSTM  & 90.83\%\\
Deep CNN    & 91.90\%\\
Deep CNN-lp    & \textbf{95.27\%}\\
\hline
\end{tabular}
\end{center}
\end{table}

\begin{figure}[htbp]
\centerline{\includegraphics[scale=1.]{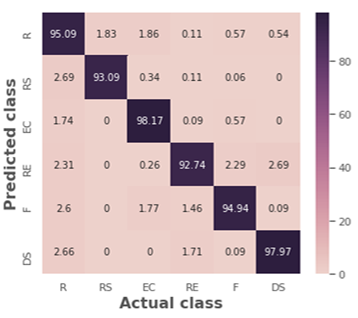}}
\caption{Plots displaying: The confusion matrix of the deep CNN-lp.
\label{fig7}}
\end{figure}

Figure~\ref{fig7} presents the confusion matrix obtained using deep CNN-lp. It is observable that around 5\% of the performed GoIs were misclassified into Random (R), probably due to the fact that many arm movements that were close to them were added to the dataset to increase the model’s robustness to false positives. Furthermore, ~2.69\% of the Recommend Evacuation (RE) gestures were classified as distress signals and almost 2\% of Fire (F) indications were falsely identified as RE and Emergency Contained (EC). It should be noted that the EC gesture produced the best F1-score (98.21\%).

\begin{figure}[htbp]
\centerline{\includegraphics[width=\columnwidth]{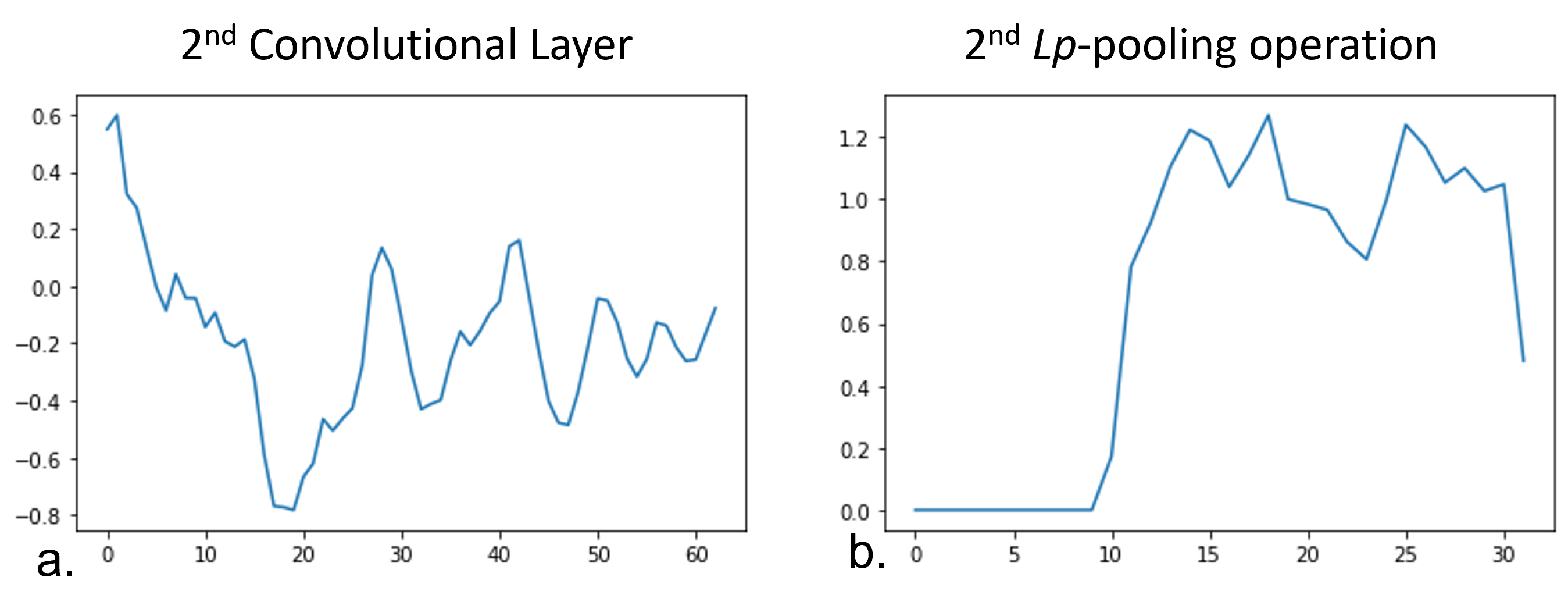}}
\caption{Visualization of the accelerometer X-axis outputs produced by the 1st kernel map of the 2nd convolutional layer (a) and the 2nd latent pooling operation (b). The dislayed signal presents a recommended evacuation gesture.\label{fig5}}
\end{figure}

Figure~\ref{fig5} displays the property of latent pooling towards paying more attention on the most import feature, and, more specifically an example of this effect that takes place on the X-axis of a recommended evacuation signal. It is observable that the values produced in the first 1.5 seconds where the subject lifted his/her arm are set to zero (i.e., they are discarded), while emphasis is given the last seconds that include the back and forth arm movement.

One of the drawbacks of the proposed approach is its size against the other evaluated methods. The latent pooling increases the number of parameters (~2,000 more parameters) and consequently its latency. More specific, inference time is around 68 ms on FOSSIL GEN 5 smartwatch, while the max pooling approach around 63 ms. Furthermore, having a closer look at the confusion matrix and projecting its results on an one hour SaR operation it leads to around 176 false positives. Thus, an error analysis based on ROC and AUC curves should be done to alter the GoI detection thresholds.

\section{Conclusion} 
One of the major issues for FR is their safety, for both the victims and the FRs who operate inside the disaster area. Arm gesture recognition technology with the use of smartwatch can improve remote communications and coordination of FR teams. To this end, we developed MoRSE smartwatch application that exploits a DL model trained over a collected dataset of arm gestures used in airport ground control operations. The proposed deep CNN architecture was evaluated against several DL and ML algorithms, obtaining accuracy over 95\%.

Future steps include MoRSE evaluation in real-world SaR operations. Furthermore, model optimization techniques (e.g., quantization or knowledge distillation \cite{Touvron2021TrainingDI}) could be applied to decrease the model's inference time,  while modality-wise relational reasoning \cite{Kasnesis2021ModalitywiseRR} will be examined for more effective sensor fusion. Finally, a usability study provided by FRs should also be helpful to improve the smartwatch application.

\section*{Acknowledgment}

This research was funded by the European Commission’s H2020 programme under project FASTER, grant number 833507.

\bibliographystyle{IEEEtran}
\bibliography{bib}

\end{document}